\documentclass{llncs}
\usepackage{algorithmic, algorithm, graphicx, subcaption, url}
 \urldef{\mailsa}\path|j.tang@adfa.edu.au,Eleni.Petraki@canberra.edu.au,h.abbass@adfa.edu.au|

\begin{document}
\title{Shaping Influence and Influencing Shaping: A Computational Red Teaming Trust-based Swarm Intelligence Model}

\author{Jiangjun Tang\inst{1} \and Eleni Petraki\inst{2} \and Hussein Abbass\inst{1}}

\authorrunning{Tang, Petraki, \& Abbass}

\institute{
 University of New South Wales, School of Engineering and Information Technology, Canberra, ACT 2600, Australia\footnote{Portions of this work was funded by the Australian Research Council Discovery
Grant number DP140102590.}.\\
 \and
 Faculty of Science, Technology, Education, and Mathematics, University of Canberra Canberra, Australia.\\
 \mailsa
  \url{}}

\toctitle{Lecture Notes in Computer Science} \tocauthor{}

\maketitle

\begin{abstract}
Sociotechnical systems are complex systems, where nonlinear
interaction among different players can obscure causal
relationships. The absence of mechanisms to help us understand how
to create a change in the system makes it hard to manage these
systems.

Influencing and shaping are social operators acting on
sociotechnical systems to design a change. However, the two
operators are usually discussed in an ad-hoc manner, without
proper guiding models and metrics which assist in adopting these
models successfully. Moreover, both social operators rely on
accurate understanding of the concept of trust. Without such
understanding, neither of these operators can create the required
level to create a change in a desirable direction.

In this paper, we define these concepts in a concise manner
suitable for modelling the concepts and understanding their
dynamics. We then introduce a model for influencing and shaping
and use Computational Red Teaming principles to design and
demonstrate how this model operates. We validate the results
computationally through a simulation environment to show social
influencing and shaping in an artificial society.


\end{abstract}

\section{Introduction}

Recently, computational social scientists are attracted to
studying means for measuring the concepts of influence and
shaping. For influence to work is to exert a form of social power.
Servi and Elson \cite{servi2014mathematical} introduce a new
definition of influence which they apply to online contexts as
`the capacity to shift the patterns of emotion levels expressed by
social media users'. They propose that measuring influence entails
identifying shifts in users' emotion levels followed by the
examination of the extent to which these shifts can be connected
with a user.
However, if the process of influence creates shifts in patterns of
emotions which can be detected in the short-term, can a persistent
application of influencing operators create a long-term shift
({\it i.e.} shaping)?

Shmueli et.al. \cite{shmueli2014sensing} discuss computational
tools to measure processes for shaping and affecting human
behaviour in real life scenarios. Trust was identified as a means
to influence humans in a social system. Moreover, trust was found
to have a significant impact on social persuasion. Trust is a
complex psychological and social concept. A review of the concept
can be found in~\cite{petraki2014trust0}.

Larson et.al. \cite{larson2009foundations} are among a few to
imply a distinction between influence and shaping, whereby shaping
is perceived to be a change to the organization or the
environment, while influence fosters attitudes, behaviours or
decisions of individuals or groups. However, the majority of the
literature follows a tendency to assume that social influencing
would lead to shaping.

In this paper, we aim to distil subtle differences to distinguish
between the two concepts. This distinction is very important for a
number of reasons. First, it examines the validity of the implicit
assumption that influencing is a sufficient condition for shaping.
Second, it is important when studying social sciences using
computational models (computational social sciences) to create
models that are not ambiguous about the social and psychological
phenomena under investigation. Third, it is vital to make it
explicit that social influencing and shaping work on different
time scales; that is, social influencing is effective in the short
run, while shaping requires time and is more effective in the long
run.

We will use a computational red teaming (CRT) model, whereby a red
agent acts on a blue team to influence, shape and sometimes
distract the blue team. The idea of the model should not be seen
from a competition or conflict perspective. The model is general,
were the red agent can be an agent that promotes a positive
attitude within a team (a servant leader) or a social worker
correcting the social attitude of a gang.

\section{Influence, Shaping, and Trust}

Influence will be defined in this paper as: an operation which
causes a short-term effect in the attitude or behaviour of an
individual, group or an organization. Shaping, on the other hand,
is defined as: an operation which causes a long-term effect in the
attitude or behaviour of an individual, group or an organization.

We use the more accurate term ``effect" rather than the term
``change" because sometimes social influence and shaping need to
operate to maintain the status quo. If agent $A$ is attempting to
influence agent $B$ by changing $B$'s behaviour, agent $C$ can
attempt to counteract agent $A$'s influence by influencing $B$ to
maintain its behaviour. Therefore, influence does not necessarily
require a change to occur.

In a strict mathematical sense, social influence would change the
parameters of a model, while social shaping would alter the
constraint system.

To illustrate the difference, we will use a model, whereby a group
of blue agents attempts to follow a blue leader. A red agent has a
self-interest to influence or shape the blue team. All agents are
connected through a network. Each agent, excluding the blue leader
and the red agent, attempts to align their behaviour with their
neighbours (the other agents it is connected to). The blue leader
attempts to reach the blue goal (a position in space). When all
agents fully trust each other, and in the absence of the red
agent's effect, it is expected that the intention of the blue
leader will propagate throughout the network. Over time, the blue
agents will also move towards the blue goal.

The red agent has a different goal. It aligns with the agents it
is connected to, but it also attempts to influence and/or shape
them towards its own goal (or away from the blue's goal). Social
influence by red is represented through changing red movements;
thus, affecting the movements of its neighbours. Social shaping by
red is represented through a network re-wiring mechanism.
Connections in a network are the constraints on the network's
topology. By rewiring the network, the red agent changes the
constraints system. We abstract trust to a scale between -1 and 1,
whereby ``1" implies maximum trust, while ``-1" implies maximum
distrust. We do not differentiate in this paper between distrust
and mistrust. A ``0" value is a neutral indicator that is
equivalent to not knowing a person.

\section{The Model}
An agent-based Boids~\cite{reynolds1987flocks} model is proposed
in this paper. All agents are randomly initialized with random
headings and locations. Agents are connected through a network
structure that allows information exchange among the agents. In
this setup, the neighborhood is mostly defined by the hamming
distance between two agents in the network, while sometimes it
will be defined by the proximity of one agent to another in the
physical space. This latter definition is the classic and default
one used in the Boids model. Three Boids rules: cohesion,
alignment, and separation, are still adopted here. However, the
first two vectors are sensed by network connections while the
separation vector is perceived through the Euclidean distance in
the physical space. Each agent has a trust factor value which
decides how much this agent trusts the information perceived from
others. The first two vectors are scaled using the trust factor
before an agent's velocity gets updated. An agent 100\% trusts the
cohesion and alignment information from its linked neighbours when
it has a trust factor of 1. When the trust factor is -1, the agent
totally believe that the information is deliberately altered to
the opposite value, and therefore, the agent reverses the
information it receives.

In the model, there are three types of agents: blue leader
($A_B$), red agent ($A_R$), and blue agent. The blue leader always
moves towards a specific location/goal, and attempts to make the
other blue agents follow it. The blue agent is an agent that
senses its neighbours through  network links for both cohesion and
alignment but by Euclidean distance for separation, and then makes
decisions on its new velocity. The red agent is a special agent in
the model who controls the level of noise ($\eta$) in the velocity
and network connections for influencing and shaping.  Many blue
agents can exist but there is only a single blue leader and a
single red agent.

Agents form a set $A$ and live in a space ($S$) defined by a given
width ($spaceW$) and a given length ($spaceL$). All agents are
connected by a random network. To establish network connections, a
probability ($p$) is defined. If we have $n$ agents including one
blue leader, one red agent, and $n-2$ blue agents, the network can
be denoted as $G(n, p)$. A Goal ($G$) is a 2-D position that sits
at one of the corners of $S$. Blue leader always aims to move
towards $G$.  The area surrounding of $G$ is denoted by $\delta$.
Once the blue leader enters this area, the position of $G$
changes. An agent has the following common attributes:

\begin{itemize}
\item Position ($p$), $p\in S$, is a 2-D coordinate.

\item Velocity ($v$) is a 2-D vector representing the agent's
movement (heading and speed) in a time unit.

\item Cohesion Velocity ($cohesionV$) of an agent is the velocity
calculated based on the mass of all agents that are connected to
this agent.

\item Alignment Velocity ($alignmentV$) of an agent is the
velocity calculated based on the average velocity of all agents
that are connected to this agent.

\item Separation Velocity ($separationV$) of an agent is the
velocity that forces this agent to keep a certain small distance
from its neighbors and is based on the Euclidean distance.

\item Velocity weights:
\begin{itemize}
\item Cohesion weight ($w_c$): a scaler for the cohesion velocity.
\item Alignment weight ($w_a$): a scaler for the alignment
velocity.
\item Separation weight ($w_s$): a scaler for the
separation velocity.
\end{itemize}

\item Trust factor ($\tau$) defines how much an agent trusts its
connected neighbours. It has an impact on both the cohesion
velocity and alignment velocity but not on the separation velocity.
\end{itemize}

All agents except the blue leader attempt moving towards the
neighbours' location guided with the cohesion vector. The cohesion
vector,$cohesionV_i$, of an agent $A_i$ is:
\begin{equation}
cohesionV_i = \frac{\sum_{j=0}^{|N|} p_j}{|N|} - p_i
\label{Equation:Cohesion}
\end{equation}
where, $|N|$ is the cardinality of the neighbourhood $N$.

The alignment velocity of an agent with its linked neighbours is:
\begin{equation}
alignmentV_i = \frac{\sum_{j=0}^{|N|} v_j}{|N|} - v_i
\label{Equation:Alignment}
\end{equation}

The separation velocity of an agent is calculated using neighbours
$N_d$ in the spatial proximity of other agents as follows:
\begin{equation}
separationV_i = -\sum_{j=0}^{|N_d|} (p_j - p_i)
\label{Equation:Separation}
\end{equation}

The trust factor of a blue agent is updated by the average trust
factors of all its connected neighbours ($N$) as below:
\begin{equation}
\tau_i = 0.5 \times (\tau_i + \frac{\sum_{j=0}^{|N|} \tau_j}{|N|})
\label{Equation:Trust}
\end{equation}
The blue leader and red agent's trust factors are not updated.

The velocity of the blue leader always aims at the goal $G$ at
each step and it is not affected by any other factor. The
velocities at time $t$ of all agents except the blue leader are
updated by Equation~\ref{Equation:Velocity}.
\begin{equation}
v = v + \tau \times ( w_c \times cohesionV + w_a \times
alignmentV) + w_s \times separationV \label{Equation:Velocity}
\end{equation}
where, $cohesionV$, $AlignmentV$, and $separationV$ are normalized
vectors. The position at time $t$ of each agent can be updated by:
\begin{equation}
p = p + v_t \label{Equation:Position}
\end{equation}

If an agent's new position is outside the bounds of $S$, the
reflection rule is applied. According to
Equation~\ref{Equation:Velocity}, an agent adjusts its own
velocity in compliance with both $cohesionV$ and $alignmentV$ when
it has a positive trust value, and follows the opposite direction
as suggested by $cohesionV$ and $alignmentV$ when its trust factor
is negative. If $\tau = 0$, only the separation vector takes
effect on this agent so that this agent doesn't anyone.

The red agent introduces heading noise, changes its network
structure, or does both at each time step. The heading noise can
be propagated to blue agents through the connections of the
network to cause a deviation in some blue agents' moving
directions. Changes in the network structure may result in long
term effects on blue agents.

At each time step, Red agent updates its own velocity
($v_{RedAgent}$) by Equation~\ref{Equation:Velocity} and then
Equation~\ref{Equation:NoiseV} uses a normal distribution
($N(0,\eta)$) to generate noise and add it to Red's velocity.
\begin{equation}
v_{RedAgent} = v_{RedAgent} + N(0,\eta)
\label{Equation:NoiseV}
\end{equation}
Equation~\ref{Equation:Position} is used to update the red agent's
position.

Furthermore, the red agent has the ability to re-configure
network connections by using the noise level $\eta$ as a
probability that governs the eventuality of the following steps:
\begin{enumerate}
 \item Randomly pick up a blue agent ($A_i$) who is connected with the red agent.
 \item Randomly pick up another blue agent ($A_j$) who is connected with $A_i$. \item Break the connection between $A_i$ and $A_j$.
 \item Connect the red agent with a randomly chosen blue agent $A_j$.
\end{enumerate}

In this way, the connection between the red agent and blue agents
changes but the number of edges of the whole network remains as
before. The long term effects of these topological updates are
expected because the path along with information propagates
changes and some blue agents may not get consistent updates from
their neighbours.

The blue leader attempts to lead other blue agents towards a given
destination, and the red agent attempts to disorient through
influence (deliberate changes in heading) and/or shaping
(deliberate re-configuration of network structure). Therefore, the
``effect'' from our model can be derived as how well the blue
agents follow the blue leader given the influence/shaping by the
red agent. A straightforward measure of this effect within our
model is the average distance between blue agents and the goal
when the blue leader reaches the goal. If this distance is small,
blue agents followed the blue leader. If it is large, red agent
distracted the blue agent.

During a single simulation run, the blue leader is tasked to reach
the goal multiple times. Each time it reaches the goal (an
iteration), the location of the goal changes. The effect is
measured at the end of each iteration. The overall effect of a
simulation run is the average of all iterations except the first
iteration, which is excluded to eliminate the warm-up period in
the system resultant from the random initialisation of agents. In
summary, the effect is defined by
Equation~\ref{Equation:DistEffect}.
\begin{equation}
\bar{d} = \frac{1}{M}\left(\sum_{m=1}^{M}{\frac{1}{n}\sum_{i=1}^n{d_{m,i}}}\right)
\label{Equation:DistEffect}
\end{equation}
where, $M$ is the number of iterations except the first one, $n$
is the number of blue agents, and $d_{m,i}$ is the distance
between agent $i$ and the goal location at the $m$'th iteration.

\section{Experimental Design}
Our aim is to evaluate the ``effects'' of the short term inference
and long term shaping caused by the red agent. Two stages are used
for the experiments. The first stage focuses on the noise of red
agent where the effect from the trust factor is minimised. The
second stage investigates both the trust factor and the red
agent's noise. The number of blue agents is 25, so there are a
total of 27 agents including a blue leader and a red agent. All
agents' initial locations are uniformly initialised at random
within $S$, where $S$ is a square space with the dimension of $500
\times 500$. All agents' headings are uniformly initialised at
random with constant speed of 1. All agents except the blue leader
have the same velocity weights: $w_c=0.4$ for cohesion, $w_a=0.4$
for alignment, and $w_s=0.2$ for separation. The initial trust
factor values of all blue agents are uniformly assigned at random
within the range of $[-1,1]$. Connections among agents are created
by a random network $G(n,0.1)$, where $n=27$.

In all experiments, two levels of noise ($\eta^-=0.1$ and
$\eta^+=0.9$) are used. In the first stage, to reduce the effect
of the trust factor, it is assumed constant with a value of 1 for
all agents; that is, all blue agents trust any perceived
information, including the information arriving from red. In the
second stage, the blue leader has two trust levels: $\tau_B^- =
0.2$ and $\tau_B^+ = 1$, and the red agent has two levels of
trust: $\tau_R^- = -0.2$ and $\tau_R^+ = -1$.

Three scenarios are designed for investigating the red agent's
impact in our experiments. In Scenario 1, the red agent introduces
noise to its heading at each time step thus this noise can
immediately affect direct neighbours and can be propagated through
the network. In Scenario 2, the red agent changes the network
structure at each time step thus shaping the environment of the
blue agents. In Scenario 3, the red agent introduces noises to its
heading and changes network structures at each time step, so that
both influence and shaping take place in our model.

Using 2\textsuperscript{k} factorial
design~\cite{montgomery2008design}, a total of 2 factor
combinations is available at the first stage and 8 at the second
stage. Each combination has three individual scenarios to study.
Moreover, the randomness exists in the initialisation phase,
therefore 10 runs for each factor combination and impact are
desired in order to obtain meaningful results. In summary, there
are 60 ($3\times2\times10$) simulation runs in the first stage and
240 ($3\times8\times10$) runs in the second stage. The results and
analysis are provided in the next section.

\section{Results and Discussion}
The results from the first stage experiments are presented in
Table~\ref{Tab:Stage1Results}. The distance between blue agents
and goal location of each run ($\bar{d}$) is listed from the
second column to the eleventh column. And the last three columns
of the table are the averages of 10 runs, the standard deviations
and the confidence intervals that are obtained at $\alpha = 0.05$.

\begin{table}
\caption{Results of red agent's noise impact when $\tau_B=1$ and $\tau_R=1$}
\label{Tab:Stage1Results}
\scriptsize
\begin{tabular}{|l|r|r|r|r|r|r|r|r|r|r|r|r|r|} \hline
&  R1 & R2 & R3 & R4 & R5 & R6 & R7 & R8 & R9 & R10 & Avg & STD & Conf. \\ \hline
\multicolumn{14}{|c|}{Scenario1: Velocity Noise} \\ \hline
$\eta = 0.1$ & 47.64 & 46.78 & 39.26 & 56.63 & 47.09 & 67.29 & 60.65 & 38.76 & 42.99 & 44.86 & 49.19 & 8.90 & 6.36 \\ \hline
$\eta = 0.9$  & 145.90 & 155.75 & 168.04 & 199.94 & 171.94 & 243.61 & 162.15 & 144.08 & 103.82 & 117.94 & 161.32 & 37.65 & 26.93\\ \hline
$e_{\eta}$ & 98.27 & 108.97 & 128.78 & 143.31 & 124.85 & 176.33 & 101.50 & 105.33 & 60.83 & 73.08 & 112.12 & 31.70 & 22.68\\ \hline
\multicolumn{14}{|c|}{Scenario 2: Network Changes} \\ \hline
$\eta = 0.1$ & 45.71 & 59.28 & 47.39 & 54.31 & 58.14 & 69.65 & 50.27 & 44.35 & 43.90 & 48.83 & 52.18 & 7.78 & 5.56\\ \hline
$\eta = 0.9$ & 61.23 & 57.63 & 56.30 & 81.25 & 53.65 & 74.69 & 55.76 & 40.86 & 47.74 & 52.03 & 58.11 & 11.36 & 8.13\\ \hline
$e_{\eta}$ & 15.52 & -1.65 & 8.91 & 26.94 & -4.49 & 5.04 & 5.49 & -3.49 & 3.85 & 3.20 & 5.93 & 9.00 & 6.44\\ \hline
\multicolumn{14}{|c|}{Scenario 3: Velocity Noise and Network Changes} \\ \hline
$\eta = 0.1$ & 45.34 & 47.09 & 65.90 & 54.05 & 51.93 & 84.91 & 54.66 & 41.11 & 43.88 & 52.21 & 54.11 & 12.23 & 8.75\\ \hline
$\eta = 0.9$  & 213.49 & 168.69 & 197.52 & 188.80 & 171.62 & 236.93 & 174.46 & 183.98 & 84.95 & 122.82 & 174.32 & 41.20 & 29.47\\ \hline
$e_{\eta}$ & 168.15 & 121.59 & 131.62 & 134.75 & 119.69 & 152.02 & 119.80 & 142.87 & 41.07 & 70.61 & 120.22 & 35.90 & 25.68\\ \hline
\end{tabular}

\end{table}

The results show that the more noise the red agent has in its
velocity, the more deviation from the goal observed by the blue
agents. Changes in the network structure can lower blue agents
performance, although the magnitude of this decrease may not be
significant. This is expected since the shaping operates work on a
smaller timescale than the influence operator. When both influence
and shaping work together, the effect is more profound than any of
the individual cases in isolation.

\begin{table}
\caption{Results of effects from red agent's noise impact and
trust factors. The confidence level is at 0.05.}
\label{Tab:Stage2Results}
\scriptsize
\begin{tabular}{|l|r|r|r|r|r|r|r|r|r|r|r|r|r|} \hline
Effect&  R1 & R2 & R3 & R4 & R5 & R6 & R7 & R8 & R9 & R10 & Avg & STD & Conf. \\ \hline
& \multicolumn{13}{|c|}{Scenario 1: Velocity Noise} \\ \hline
$e_{\tau_B}$ & -170.40 & -146.54 & -83.92 & -55.00 & -131.83 & -6.05 & -128.09 & -110.66 & -184.81 & -152.86 & -117.02 & 55.01 & 39.35\\ \hline
$e_{\tau_R}$ & 165.32 & 160.08 & 95.83 & 71.21 & 149.41 & 8.11 & 133.20 & 111.97 & 167.79 & 160.23 & 122.31 & 51.75 & 37.02\\ \hline
$e_N$ & 1.78 & 15.84 & -9.18 & 6.22 & 6.74 & 3.40 & -13.50 & 0.35 & -14.64 & -17.58 & -2.06 & 11.04 & 7.90\\ \hline
& \multicolumn{13}{|c|}{Scenario 2: Network Changes} \\ \hline
$e_{\tau_B}$ & -122.99 & -165.55 & -144.34 & -64.94 & -168.15 & -8.09 & -154.27 & -170.61 & -189.59 & -187.66 & -137.62 & 58.31 & 41.71\\ \hline
$e_{\tau_R}$ & 142.72 & 177.17 & 154.10 & 77.45 & 186.62 & 24.03 & 164.25 & 172.21 & 171.13 & 172.02 & 144.17 & 52.25 & 37.38\\ \hline
$e_N$ & 42.90 & 1.75 & 12.81 & 19.08 & 8.50 & -1.08 & -0.29 & -12.17 & 25.41 & -15.03 & 8.19 & 17.62 & 12.61 \\ \hline
& \multicolumn{13}{|c|}{Scenario 3: Velocity Noise and Network Changes} \\ \hline
$e_{\tau_B}$ & -151.65 & -140.59 & -132.44 & -35.97 & -166.98 & -7.63 & -159.37 & -171.89 & -194.38 & -163.85 & -132.47 & 61.12 & 43.72\\ \hline
$e_{\tau_R}$ & 157.63 & 152.23 & 147.97 & 38.06 & 176.90 & 16.03 & 175.96 & 163.48 & 171.95 & 174.82 & 137.50 & 59.31 & 42.43\\ \hline
$e_N$ & 2.27 & 22.60 & 15.16 & 14.70 & 21.23 & 4.64 & 5.73 & 10.22 & 20.50 & 8.15 & 12.52 & 7.38 & 5.28\\ \hline
\end{tabular}

\end{table}

The results from the second stage are summarised in
Table~\ref{Tab:Stage2Results}. Interestingly, the trust factors of
the blue leader and red agent become critical but the red agent's
noise is not critical. The model responses to the trust factor as
expected. When the blue leader has a higher level of trust, all
blue agents follow it better (smaller effect values can be
observed from $e_{\tau_B}$). On the other hand, the blue agents
demonstrate disorder behaviours (larger value of $e_{\tau_R}$) if
red agents have small negative trust. These situations are found
in all three scenarios and the effects of trust factors are all
statistically significant. Although the red agent's noise has some
effects on blue agents, it is very little and can be ignored when
compared to the effect of the trust factor. Negative trust values
taken by the blue agents counteract the influence generated by
both blue and red agents.

The red agent's noise can have impact on blue agents' behaviours
through short term influence (velocity) and long term shaping
(network structures) if the effects of trust are low. When the
trust factors are high, the situation changes. Trust has a
significant impact on blue agents' behaviours.

\begin{figure}[h]
\center
\begin{subfigure}[b]{0.3\textwidth}
    \includegraphics[width=\textwidth, height = 0.8\textwidth]{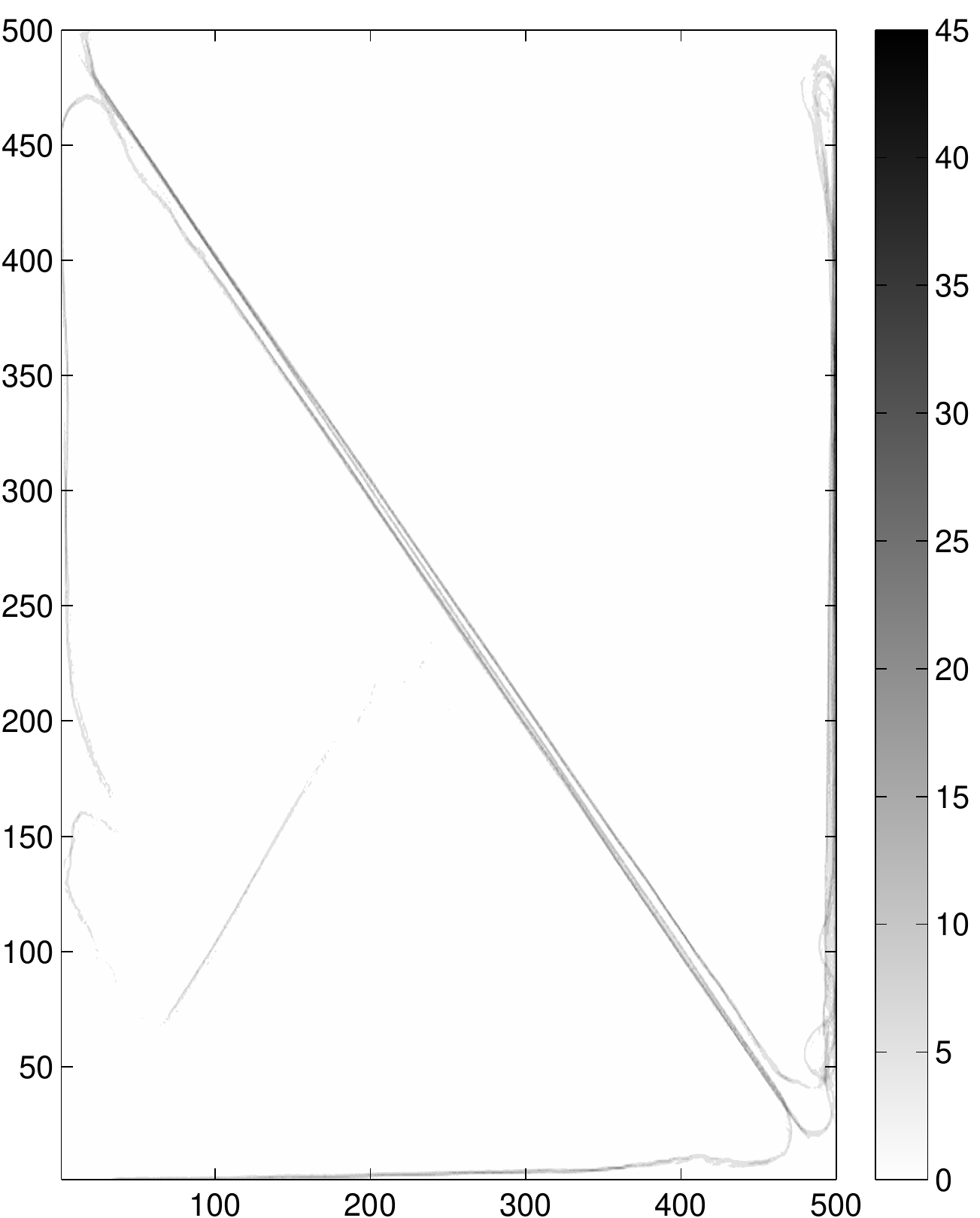}
    \caption{Scenario 1: $\eta=0.1$}
    \label{Fig:PFNoise1}
\end{subfigure}
\begin{subfigure}[b]{0.3\textwidth}
    \includegraphics[width=\textwidth, height = 0.8\textwidth]{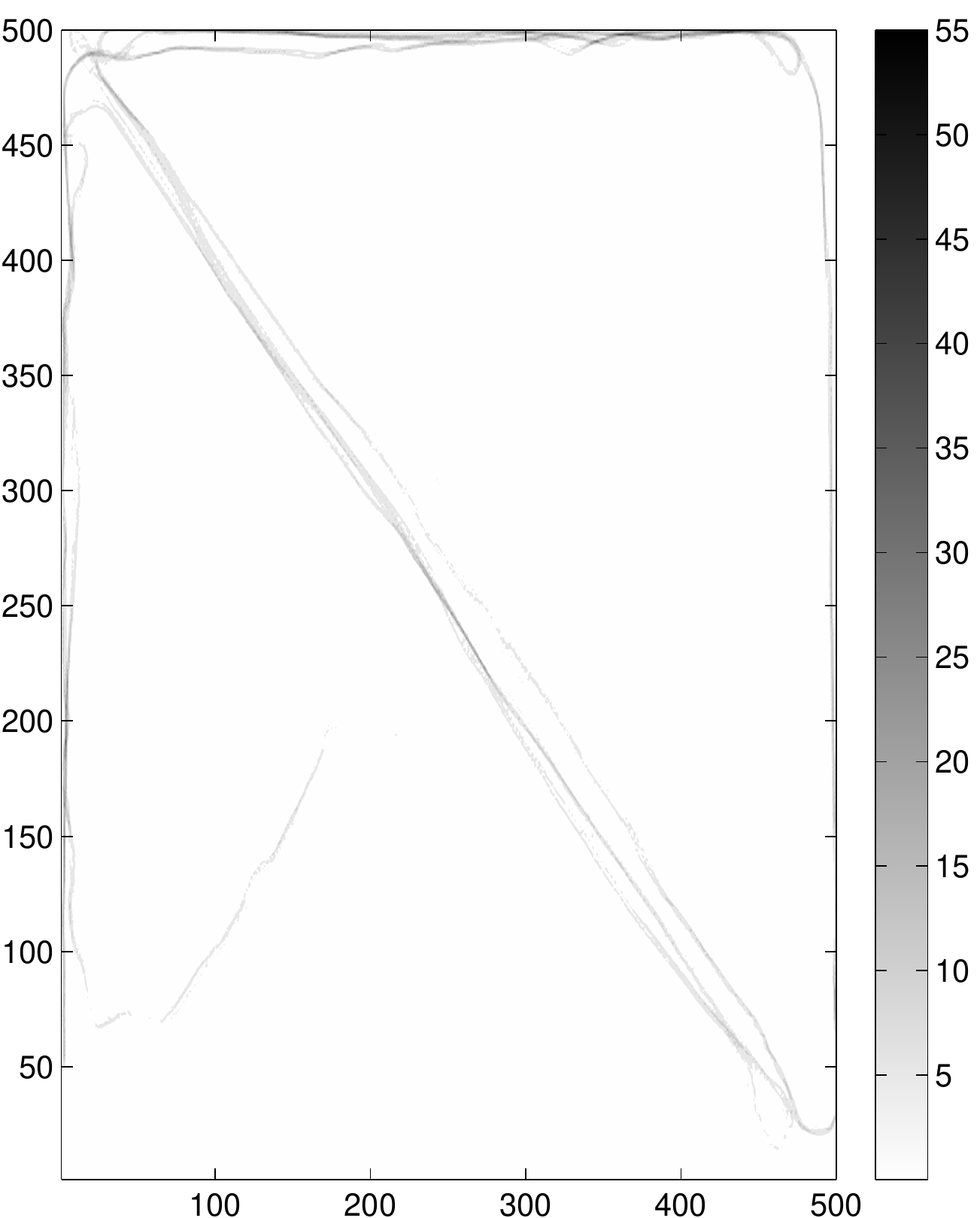}
    \caption{Scenario 2: $\eta=0.1$}
    \label{Fig:PFNoise3}
\end{subfigure}
\begin{subfigure}[b]{0.3\textwidth}
    \includegraphics[width=\textwidth, height = 0.8\textwidth]{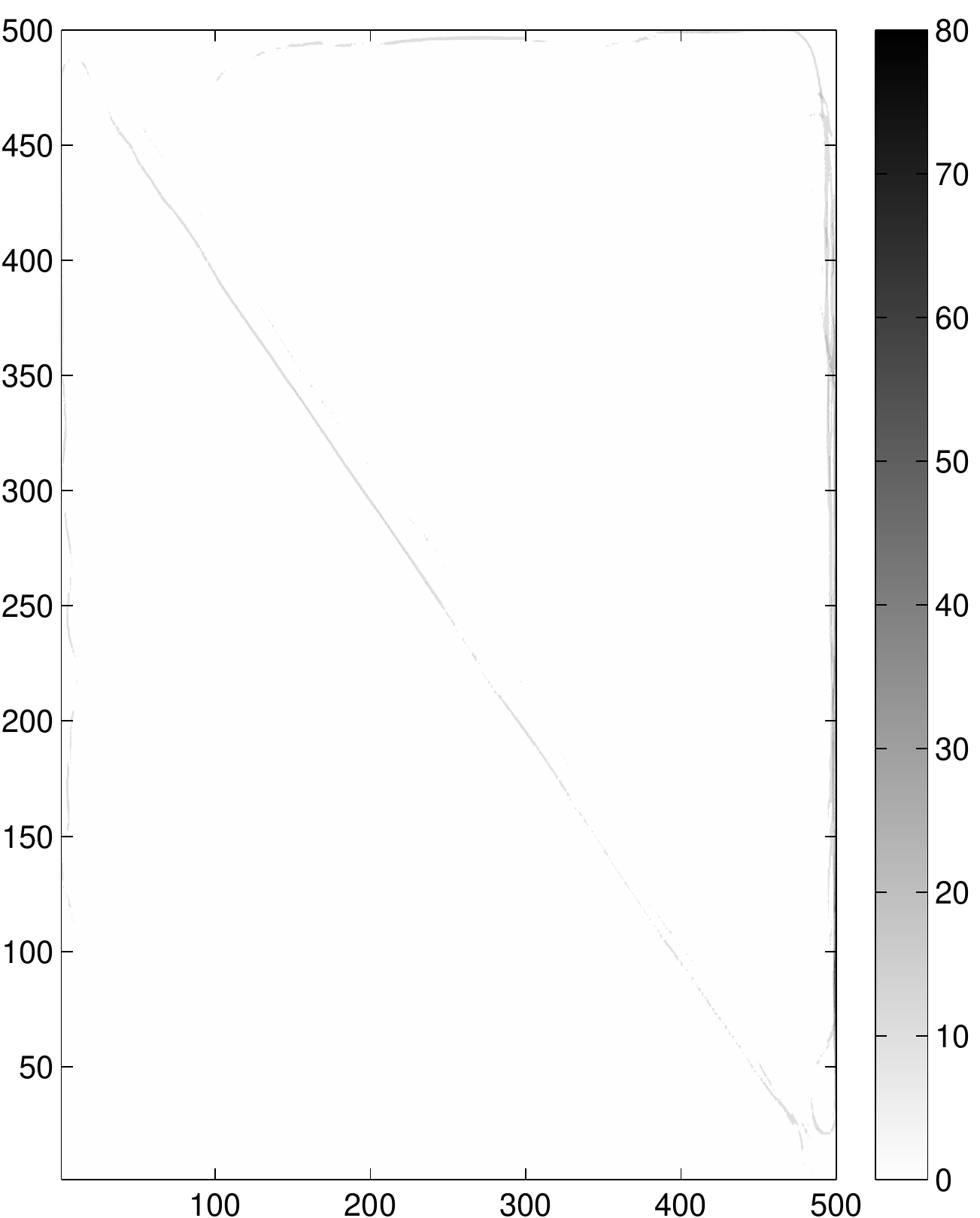}
    \caption{Scenario 3: $\eta=0.1$}
    \label{Fig:PFNoise5}
\end{subfigure}
\begin{subfigure}[b]{0.3\textwidth}
    \includegraphics[width=\textwidth, height = 0.8\textwidth]{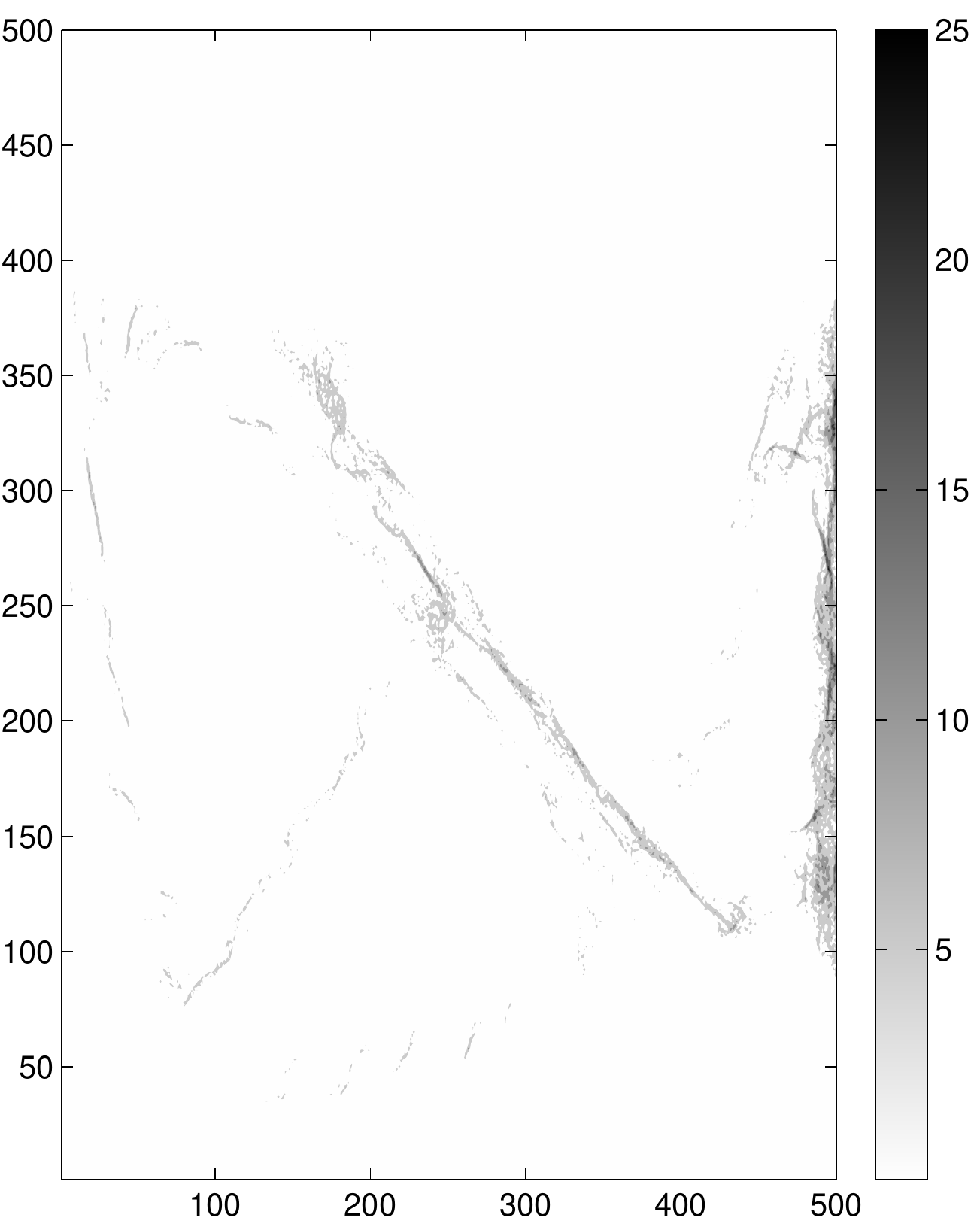}
    \caption{Scenario 1: $\eta=0.9$}
    \label{Fig:PFNoise2}
\end{subfigure}
\begin{subfigure}[b]{0.3\textwidth}
    \includegraphics[width=\textwidth, height = 0.8\textwidth]{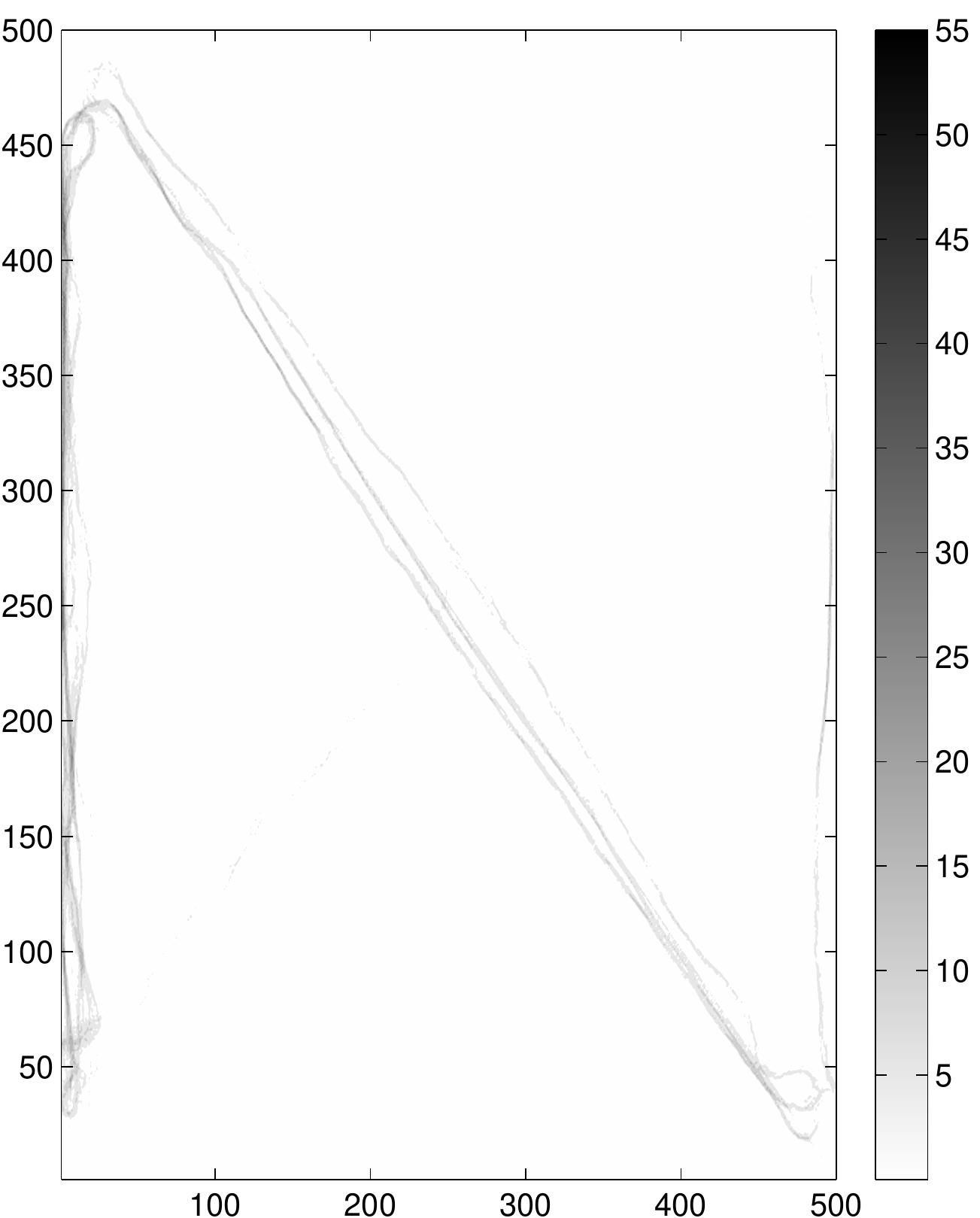}
    \caption{Scenario 2: $\eta=0.9$}
    \label{Fig:PFNoise4}
\end{subfigure}
\begin{subfigure}[b]{0.3\textwidth}
    \includegraphics[width=\textwidth, height = 0.8\textwidth]{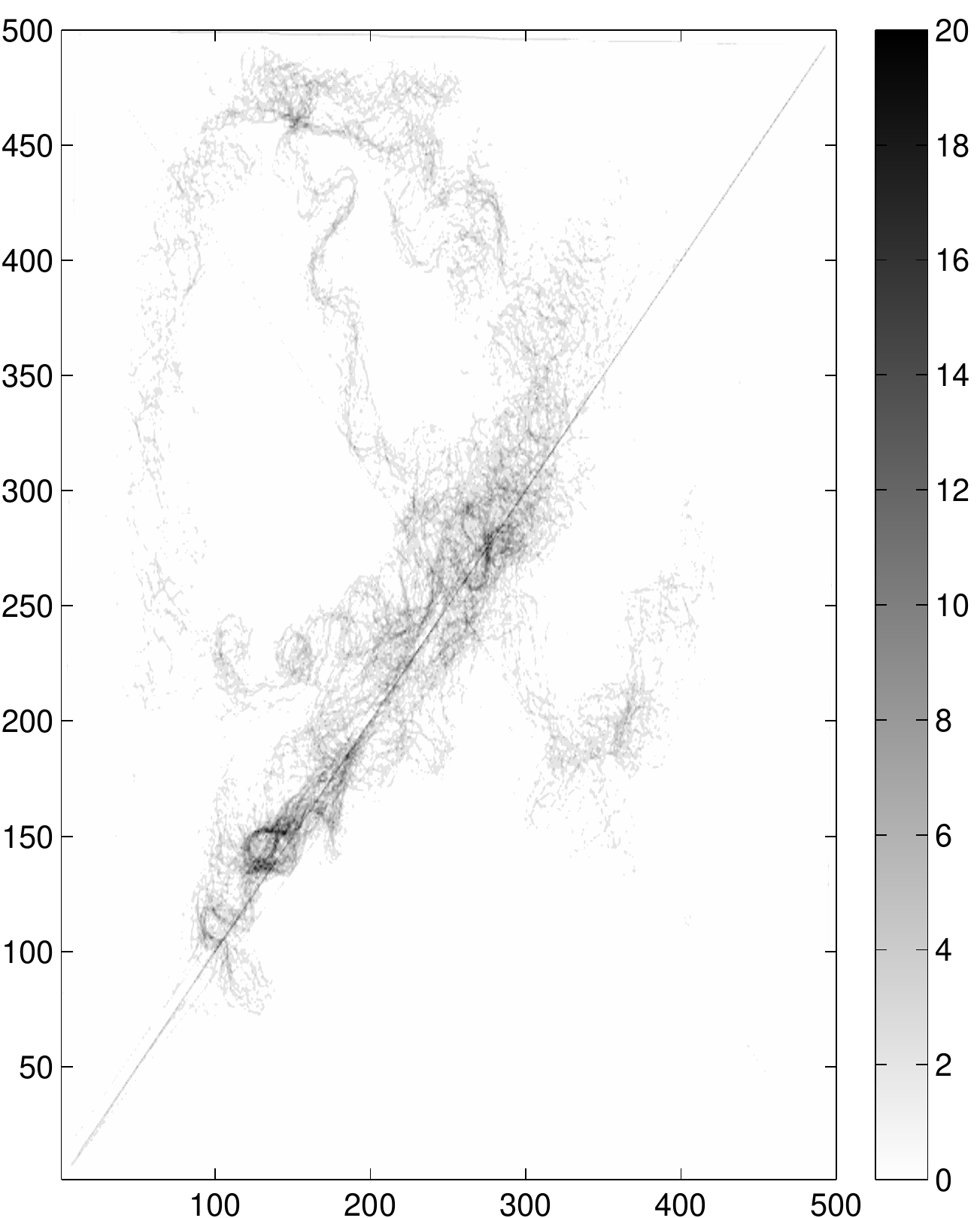}
    \caption{Scenario 3: $\eta=0.9$}
    \label{Fig:PFNoise6}
\end{subfigure}
\caption{Agents' footprints under Red agent's noise ($\eta$) impacts on velocity and network with minimum trust effects ($\tau_B=1$ and $\tau_R=1$).}
\label{Fig:FootprintsNoise}
\end{figure}

Figure~\ref{Fig:FootprintsNoise} illustrates the agents'
footprints when the red agent impacts velocity, network structure
or both but with minimum trust effects. These footprints are
obtained from the first runs of all three scenarios in the first
stage and the results are listed in the column ``R1'' of
Table~\ref{Tab:Stage1Results}.

Figure~\ref{Fig:PFNoise1}, ~\ref{Fig:PFNoise3},
and~\ref{Fig:PFNoise5} show that the blue leader leads other
agents towards the goal well as being demonstrated by a few
congested trajectories. When noise increases, blue agents'
trajectories are disturbed as shown in Figure~\ref{Fig:PFNoise2}.
Figure~\ref{Fig:PFNoise4} shows that changes in the network
structure seem to not generate much effects on blue agents'
behaviours. However, the blue agents behaviours are more random
when red affects both velocity and network structure. This
manifests disorderliness as more scattered blue agents' footprints
can be observed in the last figure.

\begin{figure}[h]
\center
\begin{subfigure}[b]{0.45\textwidth}
    \includegraphics[width=\textwidth, height = 0.8\textwidth]{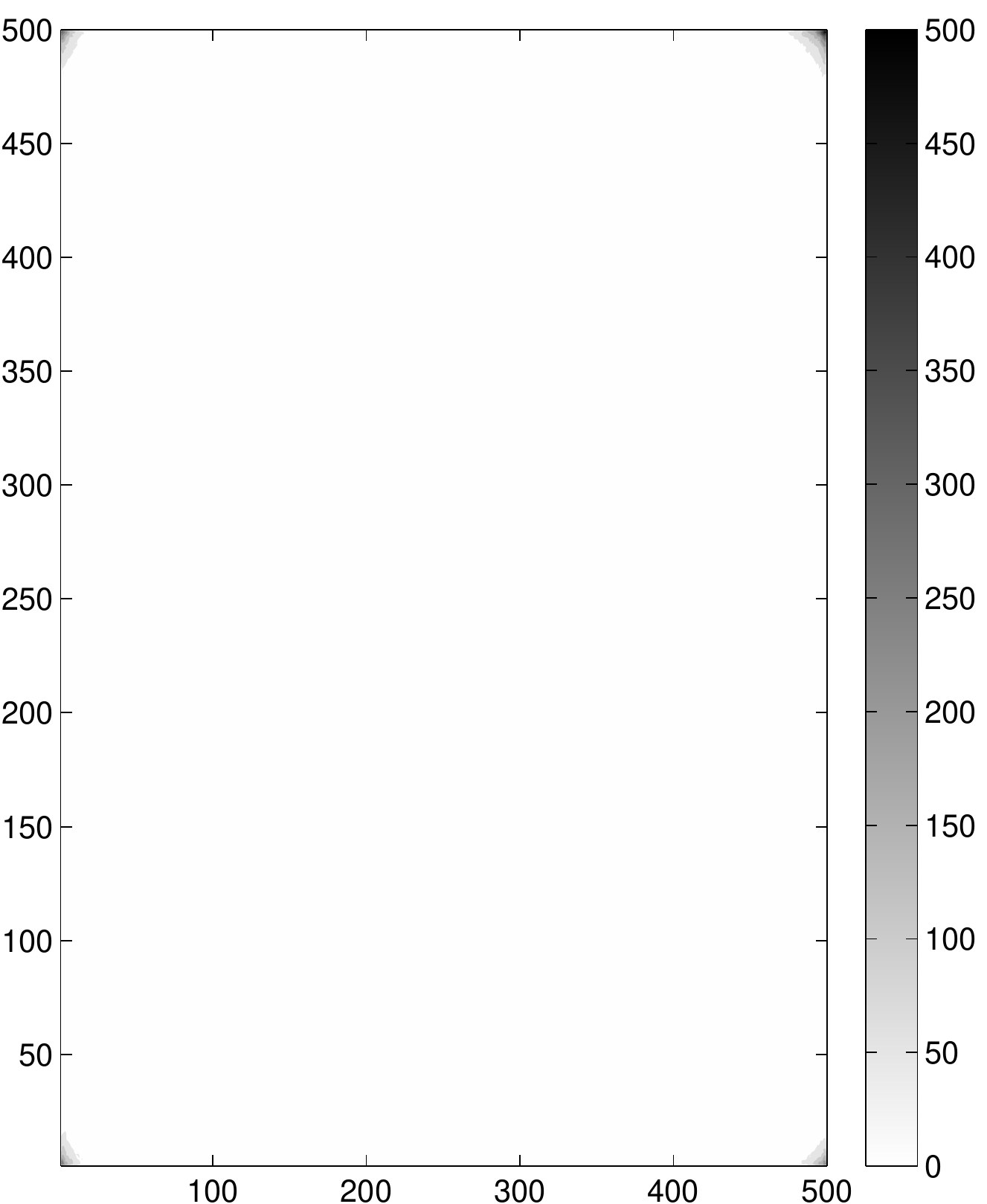}
    \caption{$\tau_B=0.2, \tau_R=-1$, and $\eta=0.1$}
    \label{Fig:PFTrust1}
\end{subfigure}
\begin{subfigure}[b]{0.45\textwidth}
    \includegraphics[width=\textwidth, height = 0.8\textwidth]{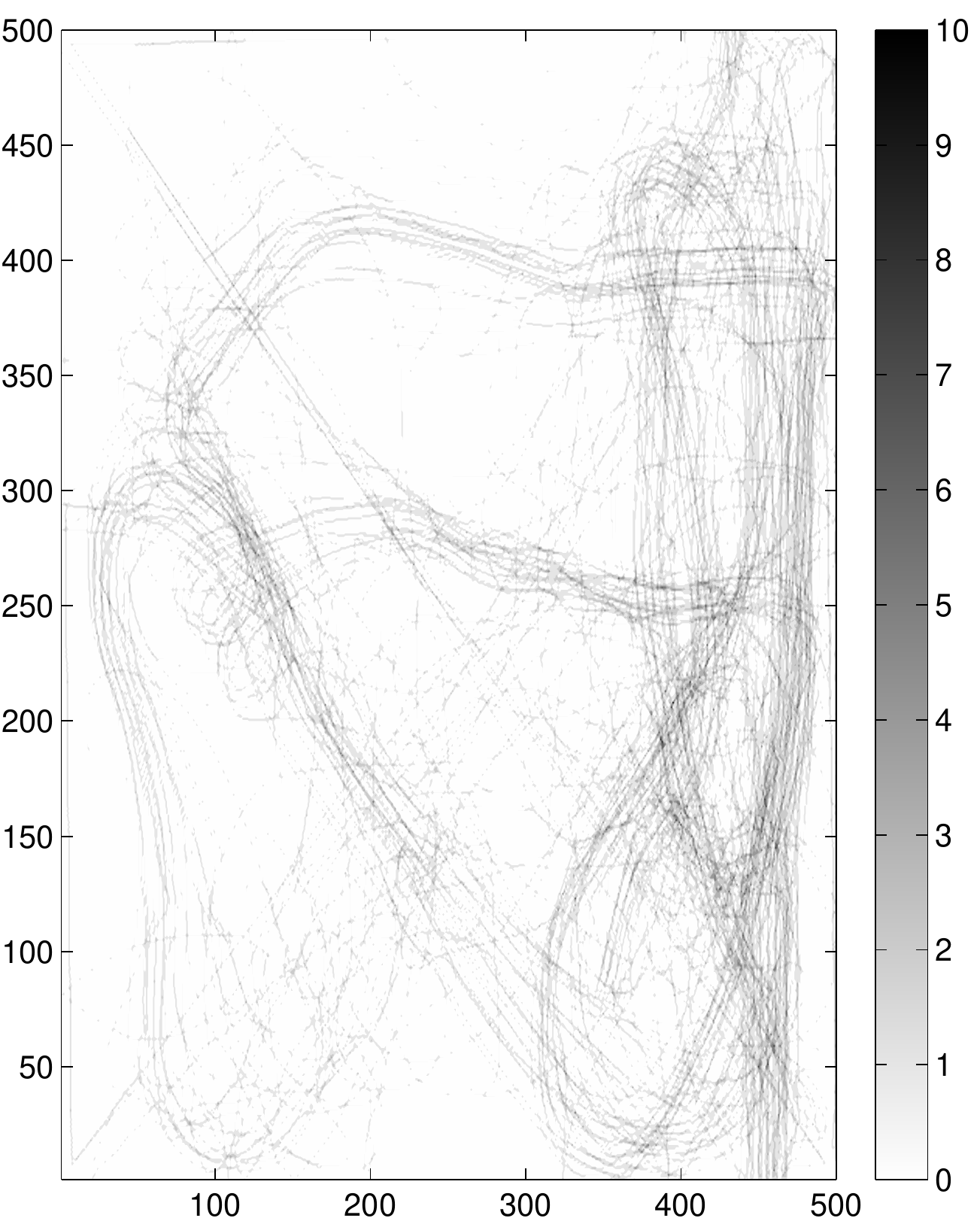}
    \caption{$\tau_B=1, \tau_R=0.2$, and $\eta=0.1$}
    \label{Fig:PFTrust2}
\end{subfigure}
\caption{Trust effects on agents behaviours with red agent noise level at 0.1 in Scenario 3.}
\label{Fig:FootprintsTrust}
\end{figure}

Figure~\ref{Fig:FootprintsTrust} shows two examples of agents'
footprints that are affected by trust with small noise values
($\eta = 0.1$). The footprints presented in
Figure~\ref{Fig:PFTrust1} are extracted from the first run of the
third scenario in the second stage with $\tau_B = 0.2$ and $\tau_R
= -1$. When the red agent's trust is -1, the negative effect on
blue agents' trust is continuously broadcasted throughout the
network. Eventually, all blue agents will have a negative trust
value that is close to -1 since the blue leader doesn't have much
power ($\tau_B=0.2$) to compete against the red agent. This
results in all blue agents distrusting each other. In this case,
the blue agents spread out to the boundaries. However, the
reflection rule forces them back into the given space, causing the
blue agents to move around the corners after several time steps as
shown in Figure~\ref{Fig:PFTrust1}.

The right side of Figure~\ref{Fig:FootprintsTrust} depicts agents'
footprints extracted from the third scenario in the second stage
with $\tau_B=1$, $\tau_R=-0.2$, and $\eta = 0.1$. Some trajectory
patterns can be observed from Figure~\ref{Fig:PFTrust2}. In this
case, the blue leader has enough power to beat the red agent in
terms of trust. All blue agents will have positive trust that are
passed from the blue leader. Although the red agent has influence
on their velocity and connections, the blue agents are still
capable to follow the blue leader to reach the goal locations
(corners) as the trajectory patterns show.

From the above examples and previous results, it can be concluded
that trust has a more significant impact on blue agents behaviours
than the effect of noise caused by the red agent.

\section{Conclusion}
In this paper, we presented a CRT trust-based model which is an
extension of the classic Boids. The network topologies for
situation awareness and a trust factor on perceived information
are introduced into our model. They provide the necessary tools to
investigate influence and shaping using CRT.

A number of experiments are designed and conducted in order to
differentiate the potential impact from influence and shaping on a
system. As the results of the first experimental stage suggest,
short term influence can have an immediate effect on the system
which is easily observed. The long term shaping effects may not be
easily observable although it has effect on the system, especially
when it interacts with influence. However, trust among agents
plays a critical role in the model. Based on our findings in the
second experiment, trust dominates the agents' behaviours
regardless of noise.

\section*{Acknowledgement}
This is a pre-print of an article published in International Conference in Swarm Intelligence, Springer, Cham, 2016. The final publication is available at \url{https://doi.org/10.1007/978-3-319-41000-5_2}.

\bibliographystyle{unsrt}

\begin{thebibliography}{1}

\bibitem{servi2014mathematical}
Les Servi and Sara~Beth Elson.
\newblock A mathematical approach to gauging influence by identifying shifts in
  the emotions of social media users.
\newblock {\em Computational Social Systems, IEEE Transactions on},
  1(4):180--190, 2014.

\bibitem{shmueli2014sensing}
Erez Shmueli, Vivek~K Singh, Bruno Lepri, and Alex Pentland.
\newblock Sensing, understanding, and shaping social behavior.
\newblock {\em Computational Social Systems, IEEE Transactions on},
  1(1):22--34, 2014.

\bibitem{petraki2014trust0}
Eleni Petraki and Hussein Abbass.
\newblock On trust and influence: A computational red teaming game theoretic
  perspective.
\newblock In {\em Computational Intelligence for Security and Defense
  Applications (CISDA), 2014 Seventh IEEE Symposium on}, pages 1--7. IEEE,
  2014.

\bibitem{larson2009foundations}
Eric~V Larson, Richard~E Darilek, Daniel Gibran, Brian Nichiporuk, Amy
  Richardson, Lowell~H Schwartz, and Cathryn~Q Thurston.
\newblock Foundations of effective influence operations: A framework for
  enhancing army capabilities.
\newblock Technical report, DTIC Document, 2009.

\bibitem{reynolds1987flocks}
Craig~W Reynolds.
\newblock Flocks, herds and schools: A distributed behavioral model.
\newblock In {\em ACM SIGGRAPH computer graphics}, volume~21, pages 25--34.
  ACM, 1987.

\bibitem{montgomery2008design}
Douglas~C Montgomery.
\newblock {\em Design and analysis of experiments}.
\newblock John Wiley \& Sons, 2008.

\end{thebibliography}

\end{document}